%% file: ArxivSubmission.tex
\documentclass[letterpaper]{article} % DO NOT CHANGE THIS
\PassOptionsToPackage{table}{xcolor} % needed for colortbl in tables
\usepackage[preprint]{aaai2027}  % DO NOT CHANGE THIS
\usepackage[hyphens]{url}  % DO NOT CHANGE THIS
\usepackage{graphicx} % DO NOT CHANGE THIS
\usepackage{natbib}  % DO NOT CHANGE THIS AND DO NOT ADD ANY OPTIONS TO IT
\usepackage{caption} % DO NOT CHANGE THIS AND DO NOT ADD ANY OPTIONS TO IT
\usepackage{algorithm}
\usepackage{algpseudocode}

\usepackage{newfloat}
\usepackage{listings}
\DeclareCaptionStyle{ruled}{labelfont=normalfont,labelsep=colon,strut=off} % DO NOT CHANGE THIS
\floatstyle{ruled}
\newfloat{listing}{tb}{lst}{}
\floatname{listing}{Listing}

\usepackage{booktabs}

\usepackage{amsmath}
\usepackage{amssymb}
\usepackage{mathtools}
\usepackage{multirow}
\usepackage{colortbl}
\usepackage{pifont}
\usepackage{soul}
\usepackage{adjustbox}
\usepackage{enumitem}

\definecolor{best}{RGB}{255,204,204}
\definecolor{second}{RGB}{255,230,153}
\definecolor{third}{RGB}{255,242,204}
\definecolor{bestcolor}{HTML}{E79696}
\newcommand{\cmark}{\ding{51}}
\newcommand{\xmark}{\ding{55}}

\providecommand{\Description}[1]{}

\title{DerainSplat: Feed-Forward Clean 3D Gaussian Splatting from Sparse Rainy Views}

\author{
    Fuzhen Jiang\textsuperscript{\rm 1}\equalcontrib,
    Changyue Shi\textsuperscript{\rm 2}\equalcontrib,
    Chuxiao Yang\textsuperscript{\rm 1},
    Xinyuan Hu\textsuperscript{\rm 1},
    Wenjie Ye\textsuperscript{\rm 1},
    Minghao Chen\textsuperscript{\rm 1}\corresponding
}
\affiliations{
    \textsuperscript{\rm 1}Hangzhou Dianzi University\\
    \textsuperscript{\rm 2}School of AI for Science, Peking University\\
}

\begin{document}

\maketitle

% =====ABSTRACT=====
\begin{abstract}
% 去雨已经有很好的发展，但是更关注在2D derain上
% 自动驾驶、具身智能要求前馈从有雨场景重建干净场景
% 但是现有方法常假设干净输入而在weather情况下fail
% 因此，我们提出derainsplat，一个前馈框架能直接从几张雨图重建干净场景
% to support this task，我们构建了一个大规模多视角去雨数据集，通过我们提出的合成雨pipeline，集成了...四阶段
% 下面方法部分一句一句传递清楚
% Extensive 实验说明derainsplat展现了...
Although image deraining has advanced substantially, existing methods mainly focus on 2D image restoration.
As spatial intelligence applications such as embodied AI and autonomous driving continue to emerge, reconstructing clean 3D scenes from sparse rainy views in a feed-forward manner becomes increasingly important.
Existing feed-forward 3D Gaussian Splatting (3DGS) methods often assume clean inputs and collapse under rainy conditions.
To this end, we present \textbf{\textit{DerainSplat}}, a feed-forward framework that reconstructs clean 3D scenes from only a few rainy views.
To support this task, we build a large-scale multi-view derain dataset through a four-stage synthesis pipeline that sequentially models overcast illumination, depth-dependent haze, rain streaks, and lens raindrops, producing privileged weather factors.
We introduce a weather net that predicts the weather factors from rainy context and yields two support maps.
Scene support modulates cross-view cost-volume matching, while radiance support drives depth-aligned appearance fusion to fill corrupted pixels.
The derived geometry evidence further attenuates Gaussian opacity to reduce spurious structures.
A rainy cycle consistency re-renders clean views using the predicted factors and aligns them with rainy inputs.
Extensive experiments show that \textbf{\textit{DerainSplat}} outperforms existing methods on various datasets, including RealEstate10K, ACID, Mip-NeRF360, and real-world rainy scenes, with strong cross-dataset generalization.
\end{abstract}


% Teaser figure (acmart's teaserfigure is not available in the AAAI style,
% so it is reproduced here as an ordinary two-column figure).
\begin{figure*}[t]
    \centering
    \includegraphics[width=\linewidth]{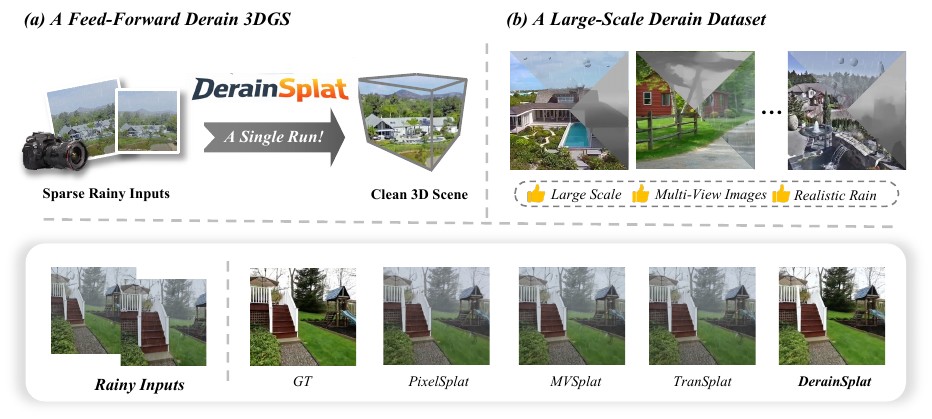}
    \vspace{-15pt}
    \caption{\textit{(a)} Given each rainy context view, \textbf{\textit{DerainSplat}} predicts a clean 3D Gaussian scene in a single forward pass. \textit{(b)} We build a large-scale multi-view derain dataset with paired clean--rain images and privileged weather factors. DerainSplat renders cleaner novel views than existing feed-forward baselines, with fewer rain/haze artifacts and stronger structural fidelity.}
    \label{fig:teaser}
    \vspace{-10pt}
\end{figure*}

% ===================MAIN===================
\input{sections/1introduction}

\input{sections/2related_work}

\input{sections/3methodology}

\input{sections/4experiment}

\input{sections/5conclusion}

\bibliography{aaai2027}


\end{document}

%% file: sections/1introduction.tex
% !TEX root = ../AnonymousSubmission2027.tex
\section{Introduction}
Image deraining has advanced substantially in recent years, evolving from supervised deep networks~\cite{zamir2021mprnet,zamir2022restormer} to unsupervised restoration with stronger real-world generalization~\cite{dong2025csud}, and from single-image deraining to joint rain-haze modeling~\cite{hu2021single}.
However, these methods focus primarily on 2D image restoration, whereas applications such as embodied AI~\cite{huang2023visual} and autonomous driving~\cite{tian2025drivingforward} ultimately require a clean 3D scene reconstruction for reliable downstream perception and decision-making.
This raises an underexplored problem: reconstructing clean 3D scenes from sparse rainy views in a feed-forward manner.

Building on 3D Gaussian Splatting (3DGS)\cite{cite:3dgs}, feed-forward methods have emerged as a promising paradigm for efficient 3D perception, predicting Gaussian scene representations from sparse multi-view images without per-scene optimization~\cite{charatan2024pixelsplat,chen2024mvsplat,zhang2025transplat}.
However, these methods are developed for clean inputs and rely heavily on reliable cross-view appearance and geometric cues for depth inference.
Once the inputs are degraded by rain and haze, the quality of multi-view matching and subsequent reconstruction deteriorates significantly, leaving a gap between the clean-input assumption and the rainy conditions faced in practice.

\begin{figure}[t]
    \centering
    \includegraphics[width=0.95\linewidth]{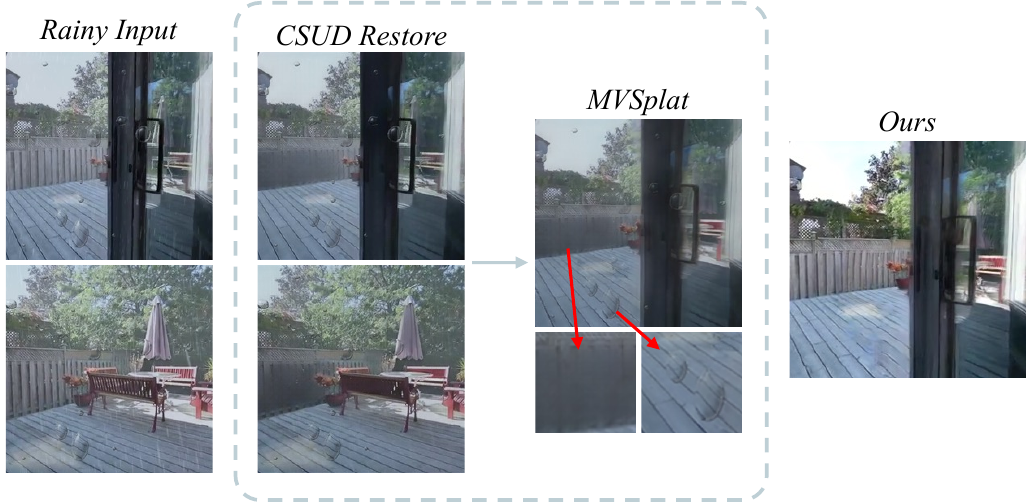}
    \vspace{-5pt}
    \caption{\textit{Failure of a two-stage strategy.} Individually plausible restored views lack cross-view consistency.}
    \label{fig:twostage_failure}
    \vspace{-17pt}
\end{figure}

A natural remedy is a two-stage pipeline that first derains each rainy input view and then reconstructs the scene with a feed-forward 3DGS model.
As shown in Fig.\ref{fig:twostage_failure}, a representative CSUD\cite{dong2025csud} + MVSplat~\cite{chen2024mvsplat} pipeline adopts this strategy, but the independently derained views often lack cross-view consistency.
Existing feed-forward 3DGS methods are not equipped to reconcile such inconsistent observations, often producing overly smooth reconstructions to accommodate the mismatched inputs across views.
These limitations suggest that rainy-to-clean 3D reconstruction should be learned directly in a multi-view, weather-aware setting rather than decomposed into independent per-view deraining followed by reconstruction.

A central bottleneck is the lack of training data that provides multi-view clean-rain supervision for feed-forward 3D reconstruction.
We build a large-scale multi-view derain dataset suite from public multi-view datasets, primarily RealEstate10K~\cite{zhou2018stereo} and ACID~\cite{liu2021infinite}, and further build a rainy Mip-NeRF360~\cite{barron2022mip} split for evaluation and real-world rainy scenes~\cite{liu2024deraings} for qualitative generalization.
Starting from clean images, we synthesize rainy observations through a structured four-stage pipeline that sequentially models global overcast illumination, depth-dependent haze, rain streaks, and lens-attached raindrops.
Besides clean-rain image pairs and camera parameters, the dataset provides privileged weather factors (transmission, airlight, rain streak layer, rain streak mask, and lens-affected mask).
These factors serve as direct supervision for the disentangled weather decomposition.

Built on this dataset, we propose \textbf{\textit{DerainSplat}}, a weather-aware feed-forward framework that directly reconstructs clean 3D Gaussian scenes from sparse rainy views.
A compact weather net first predicts disentangled weather factors from each rainy context view, supervised by the privileged factors described above.
These factors are converted into two support maps: a scene support that softly down-weights degraded regions to modulate cross-view cost-volume matching, and a radiance support that drives depth-aligned appearance fusion to fill corrupted pixels with reliable cross-view observations.
The resulting geometry evidence further attenuates Gaussian opacity at unreliable locations, reducing spurious structures without discarding valid scene content.

The clean reconstruction loss and weather supervision alone do not enforce that the rendered clean view and predicted factors jointly explain the observed rainy input.
We therefore introduce a rainy cycle consistency regularization that re-renders the clean context view with the predicted factors and aligns it with the rainy input.
Lens-affected pixels cannot be replayed by this differentiable image-formation model and are excluded via a learned censor mask, which is annealed from the ground-truth mask to the predicted one during early training.
This encourages the model to separate stable scene content from transient weather corruption.

Our contributions are summarized as follows:
\begin{itemize}
    \item \textbf{A Large-Scale Derain Dataset.} We construct a large-scale multi-view derain dataset suite through a unified four-stage synthesis pipeline, providing paired clean-rain images and privileged weather factors for supervision.
    \item \textbf{A Novel Framework.} We propose \textbf{\textit{DerainSplat}}, the first feed-forward framework that reconstructs clean 3D Gaussian scenes from sparse rainy views by disentangling weather factors into support maps that jointly guide cross-view geometry matching and appearance fusion.
    % To the best of our knowledge, this is the first work to address this task in a feed-forward manner.
    \item \textbf{Robust Performance.} Extensive experiments on various datasets demonstrate that \textbf{\textit{DerainSplat}} outperforms existing methods, with strong cross-dataset generalization.
\end{itemize}

%% file: sections/2related_work.tex
% !TEX root = ../AnonymousSubmission2027.tex
\section{Related Work}
\subsection{2D Image Deraining}
Single-image deraining has progressed from model-driven methods exploiting rain formation cues and sparse priors~\cite{garg2007visionandrain,kang2012automatic,luo2015dsc}, to deep networks trained on paired data~\cite{fu2017derainnet,yang2017jorder,zhang2018didmdn,ren2019prenet}, and recently to stronger restoration backbones~\cite{zamir2021mprnet,zamir2022restormer} and unsupervised deraining with better real-world generalization~\cite{dong2025csud}.
A rainy image is physically the superposition of a clean background and a rain streak layer, while atmospheric scattering further introduces depth-dependent haze~\cite{narasimhan2002vision}.
Recent works jointly model rain and haze to better reflect this formation process~\cite{hu2021single,wang2020rethinking}.
However, these methods target 2D restoration and overlook cross-view geometric consistency.

\subsection{Rainy Image Datasets}
Early deraining benchmarks are mostly synthetic single-image datasets that overlay rain streaks on clean images~\cite{fu2017derainnet,yang2017jorder,zhang2018didmdn}; SPANet points out that their rain shape, direction, and intensity lack realism~\cite{wang2019spanet}.
Later datasets improve realism: RainCityscapes jointly models rain streaks and depth-related haze~\cite{hu2021single}, SPA-Data and GT-RAIN are built from real rainy videos or controlled captures~\cite{wang2019spanet,ba2022gtrain}, and LHP-Rain and GTAV-NightRain further expand scale and photometric realism~\cite{lhprain2024,gtavnightrain2022}.
However, datasets above target single-image restoration and provide neither synchronized multi-view observations, cross-view aligned clean--rain pairs, camera calibration, nor interpretable weather factors.

\subsection{3D Reconstruction under Adverse Weather}
NeRF~\cite{mildenhall2020nerf} and 3DGS~\cite{cite:3dgs} have advanced sparse-view 3D reconstruction, but existing methods assume clean inputs and degrade when rain and haze corrupt cross-view matching.
A line of work optimizes a clean scene representation from degraded inputs per scene: Dehazing-NeRF~\cite{cite:dehazenerf} and ScatterNeRF~\cite{cite:scatternerf} handle fog, while WeatherGS~\cite{qian2025weathergs}, DeRainGS~\cite{liu2024deraings}, RainyScape~\cite{rainyscape2024}, and REVR-GSNet~\cite{revrgsnet2025} target rain; all require per-scene optimization and do not generalize across scenes.
Feed-forward 3DGS methods such as PixelSplat~\cite{charatan2024pixelsplat}, MVSplat~\cite{chen2024mvsplat}, MVSplat360~\cite{chen2024mvsplat360}, and TranSplat~\cite{zhang2025transplat} predict Gaussian representations in a single forward pass, but are developed and evaluated under clean-image settings.
DerainSplat is the first feed-forward framework tailored for rainy sparse inputs and is accompanied by a large-scale multi-view derain dataset suite.

%% file: sections/3methodology.tex
% !TEX root = ../AnonymousSubmission2027.tex
\section{Framework of DerainSplat}

In this section, we present \textbf{\textit{DerainSplat}}, a weather-aware feed-forward framework for rainy--to--clean 3D reconstruction.
We first build a multi-view derain dataset in Sec.~3.1.
A weather net then predicts disentangled weather factors and converts them into support maps.
Based on these support maps, our proposed pipeline suppresses degraded pixels for geometry, fills them with depth-aligned neighbors for appearance, and attenuates their Gaussian opacity (Sec.~3.2).

\subsection{Derain Dataset Synthesis}
\label{sec:derain_dataset_synthesis}

\begin{figure}
    \centering
    \includegraphics[width=\linewidth]{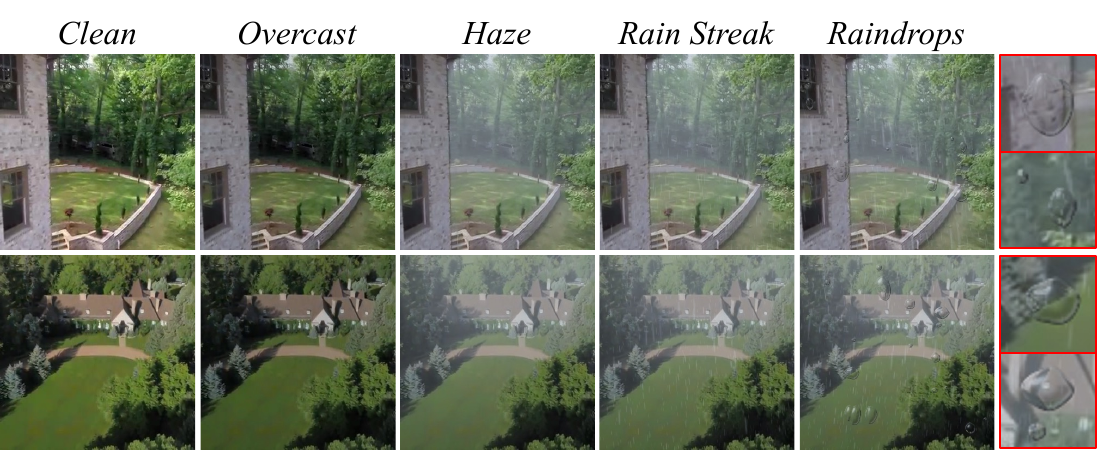}
    \caption{\textit{Stage-wise visualizations of the derain dataset synthesis process.} Starting from clean images, we sequentially synthesize overcast appearance, depth-dependent haze, rain streaks, and lens raindrops to obtain final rainy observations.}
    \label{fig:dataset_overview}
    \vspace{-15pt}
\end{figure}

To support rainy--to--clean feed-forward 3D reconstruction, we synthesize a multi-view derain dataset by decomposing rainy image formation into four sequential stages:
1) a global overcast transformation, 2) a depth-dependent haze layer, 3) a rain streak layer, and 4) a lens raindrop layer.
Fig.~\ref{fig:dataset_overview} visualizes the result of each stage.
Given a clean image $\mathbf{I}^{c}\in[0,1]^{H\times W\times 3}$, the synthesis follows:
\[
\mathbf{I}^{c}\ \rightarrow\ \mathbf{I}^{o}\ \rightarrow\ \mathbf{I}^{h}\ \rightarrow\ \mathbf{I}^{rs} \rightarrow\ \mathbf{I}^{rd},
\]
where $\mathbf{I}^{o}$, $\mathbf{I}^{h}$, $\mathbf{I}^{rs}$, and $\mathbf{I}^{rd}$ are the outputs of the overcast, haze, rain streak, and lens raindrop stages, respectively.
The complete procedure is summarized in Alg.~\ref{alg:rain_dataset_procedure}.

\begin{figure*}
  \centering
  \includegraphics[width=\linewidth]{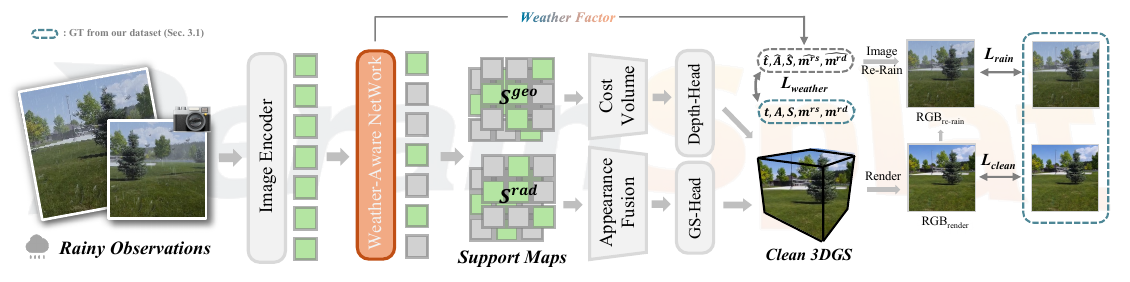}
  \vspace{-15pt}
  \caption{\textit{Overview of DerainSplat}. Rainy context views are processed by the weather net to estimate weather factors $(\hat{t}, \hat{A}, \hat{S}, \hat{m^{rs}}, \hat{m^{rs}})$ and support maps that guide cost-volume matching, apearance fusion, and opacity, yielding clean 3DGS.}
  \label{fig:pipeline}
  \vspace{-10pt}
\end{figure*}

\noindent\textbf{Global Overcast.}
Real rainy scenes exhibit scene-wide dimming and desaturation.
We quantify this with a scene-level sunny score $\gamma\in[0,1]$, estimated from the mean luminance and the bright-pixel ratio of the clean views (Alg.~\ref{alg:rain_dataset_procedure}); a larger $\gamma$ indicates a brighter, more sunlit scene.
We convert $\mathbf{I}^{c}$ to HSV space and scale its saturation and brightness by $\gamma$, so brighter scenes are dimmed and desaturated more aggressively: $\mathbf{I}^{o}=\mathcal{T}(\mathbf{I}^{c};\gamma).$
The sunny score is shared across all views of a scene to keep weather consistent.

\noindent\textbf{Depth-dependent Haze.}
We obtain the hazy image $\mathbf{I}^{h}$ using atmospheric scattering model~\cite{narasimhan2002vision}:
\begin{equation}
\label{eq:haze}
\mathbf{I}^{h}=\mathbf{I}^{o}\odot \mathbf{t}+\mathbf{A}\odot(1-\mathbf{t}),
\end{equation}
where $\mathbf{A}\in[0,1]^3$ is an airlight color and $\mathbf{t}\in[0,1]^{H\times W}$ is a pixel-wise transmission map.
We estimate a multi-view depth $\mathbf{D}$ with Depth Anything~3 (DA3)~\cite{lin2025depth}, normalize it to $\tilde{\mathbf{d}}\in[0,1]$, and set $\mathbf{t}=\exp(-\beta\tilde{\mathbf{d}})$.
Both $\gamma$ and the depth normalization range are shared across views of a scene, so the same 3D point receives consistent haze.

\noindent\textbf{Rain Streaks.}
We then overlay transient rain streaks layer on $\mathbf{I}^{h}$ to obtain $\mathbf{I}^{rs}$ via screen blending:
\begin{equation}
\label{eq:screen}
\mathbf{I}^{rs}=1-(1-\mathbf{I}^{h})\odot(1-\mathbf{S}).
\end{equation}
As detailed in Alg.~\ref{alg:rain_dataset_procedure} (Stage III), the streak layer $\mathbf{S}\in[0,1]^{H\times W\times 3}$ is built by convolving a sparse seed map with three scales of tapered, near-vertical motion kernels, then scaled by the local luminance of $\mathbf{I}^{h}$ so streaks are more visible against bright backgrounds.
We further derive a rain streak mask from $\mathbf{S}$ to mark pixels covered by visible streaks.

\begin{figure}
    \centering
    \vspace{-10pt}
    \includegraphics[width=1\linewidth]{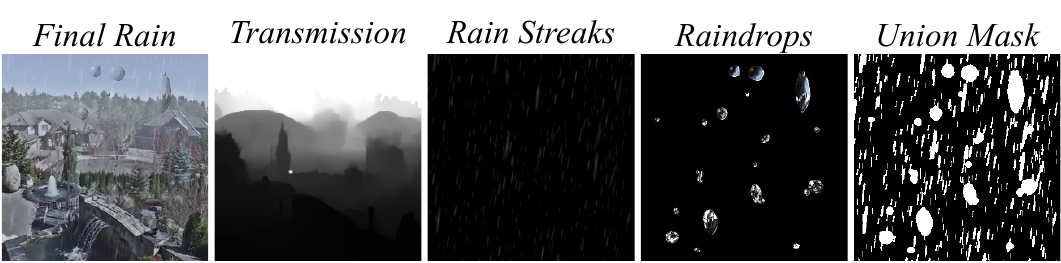}
    \vspace{-15pt}
    \caption{\textit{Visualization of weather factor.}}
    \label{fig:factor_visualization}
    \vspace{-20pt}
\end{figure}

\begin{algorithm}
    \caption{Procedure for Synthetic Derain Dataset}
    \label{alg:rain_dataset_procedure}
    \renewcommand{\algorithmicensure}{\textbf{Output:}}
    \begin{algorithmic}[1]
    \footnotesize
    \Require Clean multi-view scenes $\{(\mathcal{I}_s^c,\mathcal{C}_s)\}_{s=1}^{M}$, $\mathcal{I}_s^c=\{I_{s,v}^c\}_{v=1}^{N_s}$ and $\mathcal{C}_s=\{(K_{s,v},T_{s,v})\}_{v=1}^{N_s}$; depth estimator $\mathcal{E}$ (DA3); tone operator $\mathcal{T}(\cdot)$; airlight color $A\in[0,1]^3$.
    \Ensure $\{(I_{s,v}^c,I_{s,v}^{rd},t_{s,v},A,S_{s,v},m_{s,v}^{rs},m_{s,v}^{rd},K_{s,v},T_{s,v})\}$.
    \For{$s=1$ \textbf{to} $M$}
      \State $\gamma_s \leftarrow \operatorname{Mean}_{v}\bigl(\operatorname{SunnyScore}(I_{s,v}^c)\bigr)$
      \State $\{D_{s,v}\}_{v=1}^{N_s} \leftarrow \mathcal{E}(\{I_{s,v}^c\}_{v=1}^{N_s})$
      \State $\tilde{d}_{s,v} \leftarrow \operatorname{Norm}(D_{s,v})$
      \State $\beta_s \leftarrow 0.5+0.8\gamma_s$,\quad $t_{s,v} \leftarrow \exp(-\beta_s\tilde{d}_{s,v})$
      \State $\mathcal{L}_s \leftarrow \operatorname{LensState}(H,W)$
      \For{$v=1$ \textbf{to} $N_s$}
          \Statex \textbf{Stage I: Global Overcast}
          \State $I_{s,v}^o \leftarrow \mathcal{T}(I_{s,v}^c;\gamma_s)$
          \Statex \textbf{Stage II: Depth-Dependent Haze}
          \State $I_{s,v}^h \leftarrow I_{s,v}^o\odot t_{s,v}+A\odot(1-t_{s,v})$
          \Statex \textbf{Stage III: Rain Streak}
          \State $S_{s,v} \leftarrow \mathbf{0}\in[0,1]^{H\times W\times 3}$
          \For{$g\in\{\text{fine},\text{medium},\text{coarse}\}$}
              \State $\theta_0\sim\mathcal{U}(-8^\circ,8^\circ),\ \delta\sim\mathcal{N}(0,\sigma_g),\ \sigma_g\in\{1.3^\circ,2.2^\circ,3.0^\circ\}$
              \State $k_{s,v,g} \leftarrow \operatorname{MotionKernel}(\theta_0+\delta)$
              \State $n_{s,v} \leftarrow \operatorname{SparseSeed}(H,W)$
              \State $S_{s,v} \leftarrow 1-(1-S_{s,v})\odot(1-\operatorname{Conv}(n_{s,v},k_{s,v,g}))$
          \EndFor
          \State $S_{s,v} \leftarrow S_{s,v}\odot\operatorname{LumScale}(I_{s,v}^h)$
          \State $I_{s,v}^{rs} \leftarrow 1-(1-I_{s,v}^h)\odot(1-S_{s,v})$
          \State $m_{s,v}^{rs} \leftarrow \mathbb{I}(\max_c S_{s,v,c}>0)$
          \Statex \textbf{Stage IV: Lens Raindrop}
          \State $I_{s,v}^{rd} \leftarrow (1-\alpha_s)\odot I_{s,v}^{rs} + \alpha_s\odot\operatorname{WarpBlur}(I_{s,v}^{rs},\mathbf{f}_s,\boldsymbol{\sigma}_s)$
          \State $I_{s,v}^{rd} \leftarrow I_{s,v}^{rd} + H_s^{\mathrm{spec}} + H_s^{\mathrm{cau}}$
          \State $m_{s,v}^{rd} \leftarrow \mathbf{m}_s^{rd}$
          \State Save $\bigl(I_{s,v}^c,I_{s,v}^{rd},t_{s,v},A,S_{s,v},m_{s,v}^{rs},m_{s,v}^{rd},K_{s,v},T_{s,v}\bigr)$
      \EndFor
    \EndFor
    \end{algorithmic}
\end{algorithm}

\noindent\textbf{Lens Raindrops.}
Lens-attached water drops introduce refraction, blur, and specular highlights that the atmospheric model cannot capture.
Rather than additive noise, drops replace the underlying pixels via a coverage mask~\cite{qian2018raindrop}, with refractive displacement coming from a height--normal--flow field~\cite{porav2019icra}.
We perturb the cap boundary for irregular shapes~\cite{roser2010accv}, add a mild defocus~\cite{alletto2019iccv}, and a directional dark rim~\cite{you2016waterdrop}.
As drops stay fixed on the lens over short windows~\cite{you2013cvpr}, we generate one scene-consistent lens state $\mathcal{L}=\{\boldsymbol{\alpha},\mathbf{f},\boldsymbol{\sigma},\mathbf{H}^{\mathrm{spec}},\mathbf{H}^{\mathrm{cau}},\mathbf{m}^{rd}\}$ per scene, where $\boldsymbol{\alpha}\in[0,1]^{H\times W}$ is a per-pixel droplet coverage map, $\mathbf{f}$ a refractive flow field, $\boldsymbol{\sigma}$ a spatially-varying blur map, and $\mathbf{H}^{\mathrm{spec}},\mathbf{H}^{\mathrm{cau}}$ highlight layers.
For each view we render the droplets by warping and blurring $\mathbf{I}^{rs}$ with $\mathbf{f}$ and $\boldsymbol{\sigma}$, blending via $\boldsymbol{\alpha}$, and compositing the highlights:
\begin{equation}
\label{eq:lens_raindrop}
\mathbf{I}^{rd} = (1-\boldsymbol{\alpha})\odot\mathbf{I}^{rs} + \boldsymbol{\alpha}\odot\operatorname{WarpBlur}(\mathbf{I}^{rs},\mathbf{f},\boldsymbol{\sigma}) + \mathbf{H}^{\mathrm{spec}} + \mathbf{H}^{\mathrm{cau}}.
\end{equation}
For training stably, we further introduce a lens-affected mask $\mathbf{m}^{rd}$ marks pixels corrupted by droplet refraction or blur.

\noindent\textbf{Final Dataset Contents.}
The dataset stores clean--rain pairs $(\mathbf{I}^{c},\mathbf{I}^{rd})$, camera parameters, and privileged weather factors $(\mathbf{t},\mathbf{A},\mathbf{S},\mathbf{m}^{rs},\mathbf{m}^{rd})$, which supervise the disentangled weather decomposition in Sec.~3.2.
Intermediate images $\mathbf{I}^{o},\mathbf{I}^{h}$ and variables $\gamma,\mathbf{D},\tilde{\mathbf{d}},\beta,\theta$ are used only during generation.
Visualization exmaple is shown in Fig.~\ref{fig:factor_visualization}.

\subsection{Weather-Aware Feed-Forward 3DGS}
\label{sec:method_ff3dgs}

\begin{figure}
    \centering
    \includegraphics[width=0.9\linewidth]{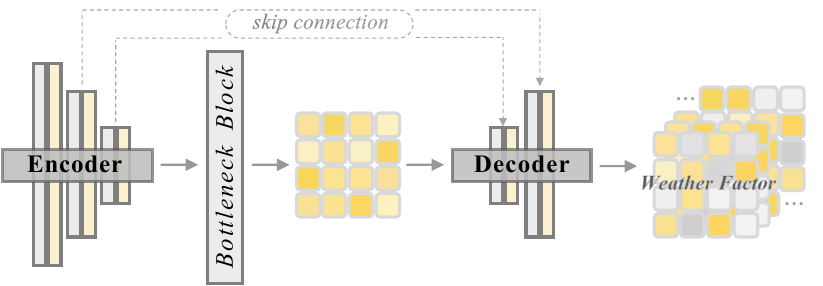}
    \caption{\textit{The architecture of the weather net.}}
    \label{fig:weather_net}
    \vspace{-18pt}
\end{figure}

\noindent\textbf{Weather Factor Prediction.}
\label{sec:method_decomp}
A compact encoder-decoder weather net (Fig.~\ref{fig:weather_net}) maps rainy context views to disentangled weather factors.
The shared trunk has three encoder blocks, a bottleneck, and two decoder blocks with skip connections; each block stacks two $3\!\times\!3$ convolutions with GroupNorm and SiLU.
A $1\!\times\!1$ head outputs pixel-wise factors---transmission $\hat{\mathbf{t}}$, rain streak layer $\hat{\mathbf{S}}$, streak mask $\hat{\mathbf{m}}^{\mathrm{rs}}$, and lens-affected mask $\hat{\mathbf{m}}^{\mathrm{rd}}$---while a global branch regresses airlight $\hat{\mathbf{A}}$ and a low-frequency color transform $(\hat{\mathbf{s}},\hat{\mathbf{b}})$.
These factors are converted into two support maps that guide the stages below.

\noindent\textbf{Support-Aware Cost Volume.}
Pixels occluded by rain or attenuated by haze yield cross-view errors.
We define a scene support that softly down-weights degraded regions:
\begin{equation}
\label{eq:scene_support}
\mathbf{s}^{\mathrm{geo}} = h(\hat{\mathbf{t}})\odot(1-\hat{\mathbf{m}}^{\mathrm{rs}})\odot(1-\hat{\mathbf{m}}^{\mathrm{rd}}),
\end{equation}
% where $h(\cdot)$ is a bounded monotonic transform of transmission.
Haze attenuates visibility gradually and is modeled by $h(\hat{\mathbf{t}})$, which retains a minimum support so distant regions still contribute geometry.
Rain is transient and directly suppress support via the masks.
Projecting $\mathbf{s}^{\mathrm{geo}}$ along epipolar lines yields a per-depth-candidate support volume that modulates the matching cost, shaping the depth posterior $\mathbf{p}$.
We summarize per-pixel geometric reliability by a geometry evidence
\begin{equation}
\label{eq:geom_evidence}
\mathbf{e} = \mathbb{E}_{\mathbf{p}}[\mathbf{s}^{\mathrm{geo}}]\odot\bigl(\eta+(1-\eta)\,\mathbf{c}\bigr),
\end{equation}
where $\mathbf{c}$ is the matching confidence and $\eta=0.1$ is a floor to avoid over-suppression.

\noindent\textbf{Depth-Aligned Appearance Fusion.}
The reference view is itself locally corrupted by weather, and naive multi-view averaging mixes clean and degraded pixels.
Using the predicted depth, we warp source pixels into the reference view to obtain depth-aligned cross-view observations.
A radiance support $\mathbf{s}^{\mathrm{rad}}=1-\hat{\mathbf{m}}^{\mathrm{rd}}$ controls a soft fusion: the reference color is retained where it is reliable, and the depth-aligned source color fills the unsupported fraction.
The fused appearance feeds the color and covariance heads.

\noindent\textbf{Geometry Evidence and Gaussian Prediction.}
The geometry evidence $\mathbf{e}$ gently attenuates Gaussian opacity via
\begin{equation}
\label{eq:opacity_evidence}
\alpha = 1-\lambda(1-\mathbf{e}),
\end{equation}
with $\lambda=0.1$, reducing spurious Gaussians at unreliable locations without discarding structure.
Based on the depth-aligned appearance and depth features, a Gaussian head regresses position, covariance, and spherical-harmonic coefficients, lifted to world space via camera.
The clean 3D Gaussians are rendered to novel viewpoints~\cite{cite:3dgs}:
\begin{equation}
\label{eq:render}
\hat{\mathbf{I}}^{c}=\mathcal{R}(\mathcal{G};\mathbf{K},\mathbf{T}).
\end{equation}

\subsection{Losses and Regularization}
\label{sec:method_losses}

We jointly optimize the weather net and the feed-forward 3DGS network with a weighted sum of three terms:
\begin{equation}
\label{eq:loss_total}
\mathcal{L}=\mathcal{L}_{\text{clean}}+\lambda_{\text{w}}\mathcal{L}_{\text{weather}}+\lambda_{\text{r}}\mathcal{L}_{\text{rain}},
\end{equation}
where $\lambda_{\text{w}}$ and $\lambda_{\text{r}}$ are balancing weights.

\noindent\textbf{Clean reconstruction loss.}
For each target view $q\in\mathcal{T}$, we render $\hat{\mathbf{I}}^{c}_{q}=\mathcal{R}(\mathcal{G};\mathbf{K}_q,\mathbf{T}_q)$ and supervise it with the clean ground view $\mathbf{I}^{c}_{q}$:
\begin{equation}
\label{eq:l_clean}
\mathcal{L}_{\text{clean}}=
\sum_{q\in\mathcal{T}}\left\|\hat{\mathbf{I}}^{c}_{q}-\mathbf{I}^{c}_{q}\right\|_2^2
+\mathbb{1}\!\left[s\geq s_{\text{p}}\right]\lambda_{\text{p}}\sum_{q\in\mathcal{T}}\mathrm{LPIPS}\!\left(\hat{\mathbf{I}}^{c}_{q},\mathbf{I}^{c}_{q}\right),
\end{equation}
where $s$ is the training step and $s_{\text{p}}=5000$. The LPIPS term is disabled early to avoid destabilizing geometry before depth has converged. This term forces the reconstructed 3D Gaussians to represent the underlying clean scene appearance.

\noindent\textbf{Weather supervision.}
The privileged weather factors $(\mathbf{t},\mathbf{A},\mathbf{S},\mathbf{m}^{\text{rs}},\mathbf{m}^{\text{rd}})$ provide direct supervision for the predicted factors:
\begin{equation}
\label{eq:l_weather}
\resizebox{\columnwidth}{!}{$\displaystyle
\mathcal{L}_{\text{weather}}=
\left\|\hat{\mathbf{t}}-\mathbf{t}\right\|_1
+\left\|\hat{\mathbf{S}}-\mathbf{S}\right\|_1
+\lambda_{\text{a}}\left\|\hat{\mathbf{A}}-\mathbf{A}\right\|_1
+\mathrm{BCE}(\hat{\mathbf{m}}^{\text{rs}},\mathbf{m}^{\text{rs}})
+\mathcal{L}_{\text{lens}}(\hat{\mathbf{m}}^{\text{rd}},\mathbf{m}^{\text{rd}}),
$}
\end{equation}
The rain-streak mask $\mathbf{m}^{\text{rs}}$ is supervised with binary cross-entropy. The lens-affected mask $\mathbf{m}^{\text{rd}}$ is sparse, so we combine three terms to prevent it from collapsing to all zeros:
\begin{equation}
\resizebox{\columnwidth}{!}{$\displaystyle
\mathcal{L}_{\text{lens}}=
\mathrm{BCE}_{\text{w}}(\hat{\mathbf{m}}^{\text{rd}},\mathbf{m}^{\text{rd}})
+\lambda_{\text{d}}\,(1-\mathrm{Dice}(\hat{\mathbf{m}}^{\text{rd}},\mathbf{m}^{\text{rd}}))
+\lambda_{\text{g}}\,\mathrm{ReLU}\!\left(\bar{\mathbf{m}}^{\text{rd}}-\bar{\mathbf{m}}^{\text{rd}}_{\text{gt}}-\delta\right)^2,
$}
\end{equation}
where $\mathrm{BCE}_{\text{w}}$ re-weights positive pixels by the negative-to-positive ratio $\lambda_{\text{d}}$, $\lambda_{\text{g}}$.
The area slack $\delta$ guard penalizes over-prediction, keeping the mask conservative. This supervision regularizes factor estimation and reliable support maps.

\noindent\textbf{Rainy cycle consistency.}
The $\mathcal{L}_{\text{clean}}$ and $\mathcal{L}_{\text{weather}}$ do not enforce that the rendered clean view and predicted factors jointly explain the rainy input. We map the rendered $\hat{\mathbf{I}}^{c}_{c}$ back to the rainy domain with the predicted factors. First, a low-frequency color transform absorbs the global overcast shift:
\begin{equation}
\hat{\mathbf{I}}^{\text{ct}}_{c}=\operatorname{clip}\!\left(\hat{\mathbf{s}}_{c}\odot\hat{\mathbf{I}}^{c}_{c}+\hat{\mathbf{b}}_{c},\,0,\,1\right),
\end{equation}
We then apply haze and screen-blend the rain streak layer:
\begin{equation}
\tilde{\mathbf{I}}^{r}_{c}=1-\left(1-\left(\hat{\mathbf{I}}^{\text{ct}}_{c}\odot\hat{\mathbf{t}}_c+\hat{\mathbf{A}}_c\odot(1-\hat{\mathbf{t}}_c)\right)\right)\odot(1-\hat{\mathbf{S}}_c).
\end{equation}

Lens refraction and blur cannot be replayed by this differentiable model, so lens-affected pixels are excluded from the loss. We align the re-rained image with the observed rainy input $\mathbf{I}^{r}_{c}$ using a robust Charbonnier penalty:
\begin{equation}
\label{eq:l_rain}
\resizebox{\columnwidth}{!}{$\displaystyle
\mathcal{L}_{\text{rain}}=\sum_{c\in\mathcal{C}}
\frac{\sum_{(i,j)}\rho\!\left(\tilde{\mathbf{I}}^{r}_{c,(i,j)}-\mathbf{I}^{r}_{c,(i,j)}\right)\,w_{c,(i,j)}}
{\sum_{(i,j)}w_{c,(i,j)}+\epsilon},
\quad
\rho(x)=\sqrt{x^2+\epsilon^2},
$}
\end{equation}
where $w_{c}=1-\mathbf{u}_{c}$ is the validity weight and $\mathbf{u}_{c}$ is the lens censor mask. The Charbonnier penalty is robust to local outliers caused by streak occlusions.

\noindent\textbf{Training stability.}
The lens censor mask is annealed from the ground truth to the predicted over initial $s_{\text{c}}=5000$ steps,
\begin{equation}
\mathbf{u}_{c}=\alpha_{s}\,\mathbf{m}^{\text{rd}}_{c}+(1-\alpha_{s})\,\hat{\mathbf{m}}^{\text{rd}}_{c},
\quad
\alpha_{s}=\max\!\left(0,\,1-s/s_{\text{c}}\right),
\end{equation}
so that the cycle loss is not disabled prematurely while the lens head is still learning. During the same window we also anneal the scene support used by the geometry path from the ground-truth-derived support toward the predicted support,
\begin{equation}
\mathbf{s}^{\text{geo}}_{c}=\alpha_{s}\,\mathbf{s}^{\text{geo,gt}}_{c}+(1-\alpha_{s})\,\mathbf{s}^{\text{geo,pred}}_{c},
\end{equation}
handing control of cross-view matching to the weather net as it becomes reliable. Finally, the support is detached from the weather net before entering the geometry path; the weather net is trained only by $\mathcal{L}_{\text{weather}}$ and $\mathcal{L}_{\text{rain}}$. This prevents the lens head from reducing the task loss by predicting an all-zero mask that retains the raw rainy reference pixels.

%% file: sections/4experiment.tex
% !TEX root = ../AnonymousSubmission2027.tex
\begin{table*}[t]
    \centering
    \scriptsize
    \renewcommand\arraystretch{1.25}
    \setlength{\tabcolsep}{3.0pt}
    \caption{\textit{Overall quantitative results on synthetic rainy benchmarks.} Feed-forward methods are evaluated on Rainy RE10K and Rainy ACID using either \textbf{official} weights or models \textbf{finetuned} on synthetic rainy data. Per-scene optimized methods are evaluated on Rainy Mip-NeRF360. All methods are evaluated on clean target views. \textbf{Bold}/\underline{underline} denote the best/second best.}
    \vspace{-5pt}
    \resizebox{\textwidth}{!}{%
    \begin{tabular}{l|ccc|ccc|ccc|l|ccc}
        \toprule
        \multicolumn{1}{c|}{\multirow{3}{*}{\textbf{Feed-Forward}}}
        & \multicolumn{3}{c|}{\textbf{Rainy RE10K (Official)}}
        & \multicolumn{3}{c|}{\textbf{Rainy ACID (Official)}}
        & \multicolumn{3}{c|}{\textbf{Rainy RE10K (Finetuned)}}
        & \multicolumn{1}{c|}{\multirow{3}{*}{\textbf{Per-Scene Optimization}}}
        & \multicolumn{3}{c}{\textbf{Mip-NeRF360}} \\
        \cmidrule(lr){2-4}\cmidrule(lr){5-7}\cmidrule(lr){8-10}\cmidrule(lr){12-14}
        & PSNR$\uparrow$ & SSIM$\uparrow$ & LPIPS$\downarrow$
        & PSNR$\uparrow$ & SSIM$\uparrow$ & LPIPS$\downarrow$
        & PSNR$\uparrow$ & SSIM$\uparrow$ & LPIPS$\downarrow$
        & 
        & PSNR$\uparrow$ & SSIM$\uparrow$ & LPIPS$\downarrow$ \\
        \midrule

        PixelSplat~\cite{charatan2024pixelsplat}
        & 17.19 & 0.662 & 0.318
        & \underline{18.80} & \underline{0.718} & \underline{0.335}
        & 17.62 & 0.683 & \underline{0.303}
        &  3DGS~\cite{cite:3dgs}
        &  \underline{16.81}
        &  0.233
        &  \underline{0.459} \\

        MVSplat~\cite{chen2024mvsplat}
        & 17.24 & 0.670 & 0.312
        & 18.74 & 0.717 & 0.342
        & 18.44 & 0.682 & 0.308
        & WeatherGS~\cite{qian2025weathergs}
        & 15.46
        & \underline{0.252}
        & 0.505 \\

        TranSplat~\cite{zhang2025transplat}
        & \underline{17.29} & \underline{0.672} & \underline{0.311}
        & 18.75 & \underline{0.718} & 0.341
        & \underline{19.04} & \underline{0.692} & 0.319
        & \textbf{Ours}
        & \textbf{19.38} & \textbf{0.524} & \textbf{0.395} \\

        \textbf{Ours}
        & \textbf{22.77} & \textbf{0.773} & \textbf{0.188}
        & \textbf{23.54} & \textbf{0.786} & \textbf{0.167}
        & \textbf{22.77} & \textbf{0.773} & \textbf{0.188}
        & \\
        \bottomrule
    \end{tabular}
    }
    \vspace{-10pt}
    \label{tab:baseline-quantitative}
\end{table*}

\begin{figure*}
    \centering
    \includegraphics[width=0.85\linewidth]{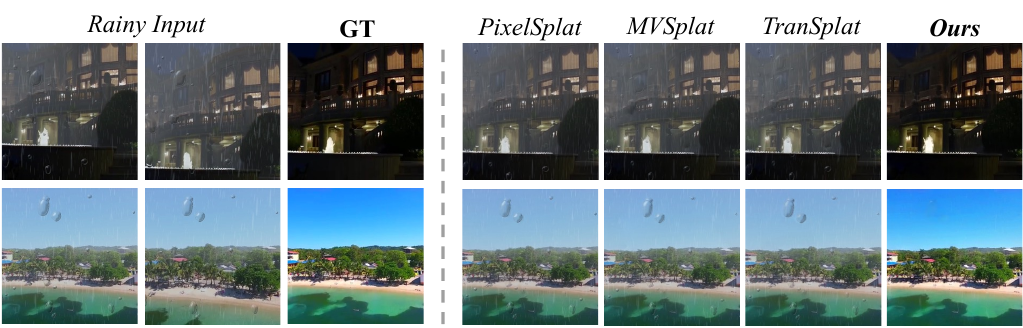}
    \vspace{-5pt}
    \caption{\textit{Qualitative comparison of novel views on rainy RealEstate10K and ACID datasets.} Compared with feed-forward baselines, DerainSplat yields cleaner novel views with fewer rain/haze artifacts and clearer scene structures.}
    \label{fig:qualitative}
    \vspace{-10pt}
\end{figure*}

\section{Experiment}
\subsection{Settings}
\noindent\textbf{Baselines.}
We compare DerainSplat against representative feed-forward 3DGS methods, including PixelSplat~\cite{charatan2024pixelsplat}, MVSplat~\cite{chen2024mvsplat}, and TranSplat~\cite{zhang2025transplat}, using their official pretrained weights for direct inference on rainy inputs.
We further compare against per-scene optimized methods, including 3DGS~\cite{cite:3dgs} and WeatherGS~\cite{qian2025weathergs}.
To ensure a fair comparison, we also retrain the feed-forward baselines from scratch on our synthetic rainy benchmarks under the same configuration as DerainSplat.

\begin{figure}
    \centering
    \includegraphics[width=0.95\linewidth]{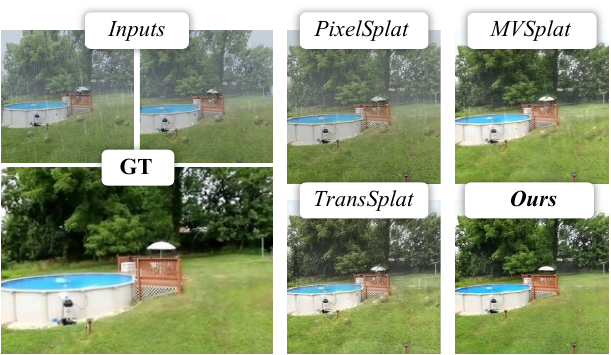}
    \vspace{-5pt}
    \caption{\textit{Finetuned qualitative comparison on rainy RE10K.}}
    \label{fig:finetuned-qualitative}
    \vspace{-18pt}
\end{figure}

\noindent\textbf{Benchmarks.}
All synthetic benchmarks are generated by the proposed four-stage synthesis pipeline, preserving the original camera parameters and clean targets.
We report PSNR, SSIM, and LPIPS for all baselines in the following sections.
Real-world rainy scenes lack aligned clean targets and are used for qualitative evaluation only.

\noindent1) \textbf{Rainy RE10K}~\cite{zhou2018stereo} contains 13,768/1,455 (train/test) rainy outdoor scenes and serves as the primary benchmark for training and in-domain evaluation.

\noindent2) \textbf{Rainy ACID}~\cite{liu2021infinite} contains approximately 11K/1.5K (train/test) rainy outdoor scenes and serves as a second large-scale in-domain benchmark.

\noindent3) \textbf{Rainy Mip-NeRF360}~\cite{barron2022mip} evaluates cross-dataset generalization and per-scene optimized methods.

\noindent4) \textbf{HydroViews}~\cite{liu2024deraings} provides real-world rainy scenes without aligned clean views for qualitative evaluation.

\noindent\textbf{Implementation Details.}
We implement \textbf{DerainSplat} in PyTorch.
Following MVSplat, both training and inference adopt a two-view setting.
Each model is trained under the default rain intensity configuration.
Training is conducted uses with a batch size of 10 across 5 NVIDIA RTX 4090 GPUs using the Adam optimizer~\cite{kingma2014adam} for 300k steps.

\begin{figure}
    \centering
    \includegraphics[width=0.79\linewidth]{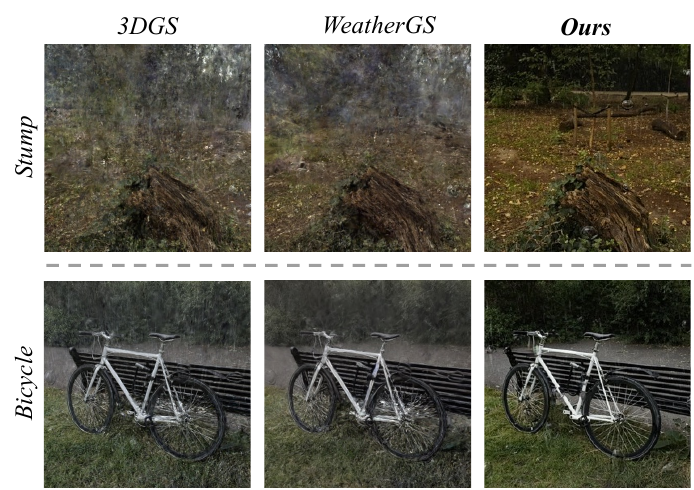}
    \vspace{-6pt}
    \caption{\textit{Qualitative comparison on rainy Mip-NeRF360.}}
    \label{fig:mipnerf-qualitative}
    \vspace{-14pt}
\end{figure}

\begin{table*}
    \centering
    \scriptsize
    \renewcommand\arraystretch{1.25}
    \setlength{\tabcolsep}{3pt}
    \caption{\textit{Analysis on two-stage derain strategy.}}
    \vspace{-5pt}
    \resizebox{0.75\textwidth}{!}{%
    \begin{tabular}{l|cccc|l|cccc}
        \toprule
        \multicolumn{1}{c|}{\multirow{3}{*}{\textbf{Feed-Forward}}}
        & \multicolumn{4}{c|}{\textbf{RealEstate10K}}
        & \multicolumn{1}{c|}{\multirow{3}{*}{\textbf{Per-Scene Optimization}}}
        & \multicolumn{4}{c}{\textbf{Mip-NeRF360}} \\
        \cmidrule(lr){2-5}\cmidrule(lr){7-10}
        & PSNR$\uparrow$ & SSIM$\uparrow$ & LPIPS$\downarrow$
        & Time$\downarrow$
        & 
        & PSNR$\uparrow$ & SSIM$\uparrow$ & LPIPS$\downarrow$
        & Time$\downarrow$
        \\
        \midrule

        MVSplat
        & 17.24 & \underline{0.670} & \underline{0.312}
        & \textbf{0.42 s}
        & 3DGS
        & \underline{16.81} &  0.233 &  0.459
        & \underline{$\approx 400$ s}
        \\

        CSUD+MVSplat
        & \underline{17.35} & 0.668 & 0.328
        & 0.59 s
        & CSUD+3DGS
        & 16.78 & \underline{0.243} & \underline{0.458}
        & $\approx 405$ s
        \\

        \textbf{Ours}
        & \textbf{22.77} & \textbf{0.773} & \textbf{0.188}
        & \underline{0.47 s}
        & \textbf{Ours} 
        & \textbf{19.38} & \textbf{0.524} & \textbf{0.395}
        & \textbf{0.47 s}
        \\
        \bottomrule
    \end{tabular}
    }
    \label{tab:twostage}
    \vspace{-3pt}
\end{table*}

\subsection{Main Results}
\noindent\textbf{Comparison with Feed-Forward Methods.}
Tab.~\ref{tab:baseline-quantitative}(left) shows that DerainSplat outperforms all feed-forward baselines on both Rainy RE10K and Rainy ACID, ranking first in all metrics with a PSNR gain of over 4 dB.
Finetuned baselines on the same rainy data closes little of the gap, attributing the advantage to the method design rather than training resources.
Fig.~\ref{fig:qualitative} and Fig.~\ref{fig:finetuned-qualitative} further show that existing methods retain rain/haze and blur structures, while DerainSplat renders cleaner novel views and more coherent structures, validating the effectiveness of weather-aware design.

\noindent\textbf{Comparison with Per-Scene Optimized Methods.}
We further evaluate DerainSplat against per-scene optimized methods on Rainy Mip-NeRF360.
Qualitative results are shown in Fig.~\ref{fig:mipnerf-qualitative} and and quantitative results are reported in Tab.~\ref{tab:baseline-quantitative}~(right).
Compared to per-scene optimized methods, DerainSplat not only achieves competitive performance but also enables real-time reconstruction.

\subsection{Other Results}
\begin{figure}
    \centering
    \includegraphics[width=0.85\linewidth]{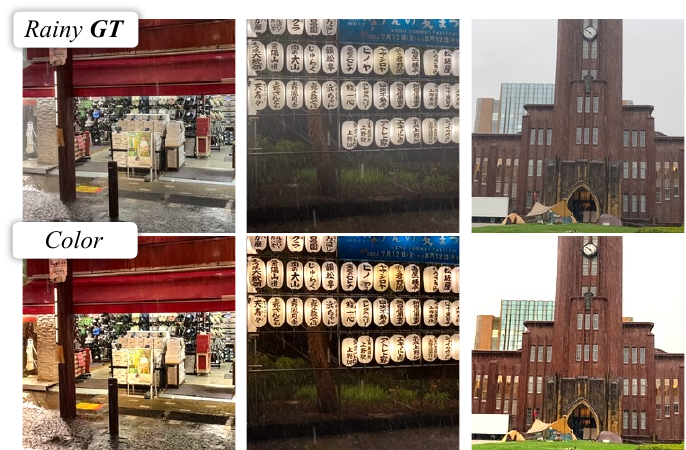}
    \vspace{-3pt}
    \caption{\textit{Qualitative results on real-world rainy scenes.}}
    \label{fig:real-qualitative}
    \vspace{-17pt}
\end{figure}

\noindent\textbf{Real-rainy Scenes Generalization.}
We evaluate DerainSplat on real rainy scenes from HydroViews~\cite{liu2024deraings} without any fine-tuning.
Since real rainy captures lack aligned clean target, we only provide qualitative results in Fig.~\ref{fig:real-qualitative}.
% DerainSplat reliably removes rain streaks and restores clear scene structure and color, with no evident geometric artifacts or domain shift, demonstrating that the rain model and deraining capability learned on synthetic rain transfer directly to real scenes.
We further validate the realism of our synthetic rain pipeline, decomposing the predicted rain-streak mask on real scenes and measuring the principal-axis angle, length, and width of each streak.
Comparison against predictions on synthetic scenes is shown in Tab.~\ref{tab:streak-geometry}, indicating that the synthetic streak geometry is physically consistent with real rainfall.

\begin{table}
    \centering
    \renewcommand\arraystretch{1.2}
    \vspace{-5pt}
    \caption{\textit{Rain-streak geometry comparison.}
    Real streaks are predicted, while synthetic sreaks
    are generated from pipeline.}
    \vspace{-5pt}
    \resizebox{0.98\linewidth}{!}{%
      \begin{tabular}{c|cc|c}
      \textit{Metric} & Real rainy (pred) & Synthetic
      (GT) & Overlap $\uparrow$\\
      \noalign{\hrule height 1pt}
      \textit{Angle} ($^\circ$) & $88.0 \pm 6.4$ &
      $89.8 \pm 6.3$ & 0.832 \\
      \textit{Length (px)} & $7.7 \pm 5.3$ & $9.4 \pm
      6.0$ & 0.842 \\
      \textit{Width (px)} & $2.59 \pm 1.68$ & $2.24
      \pm 1.49$ & 0.843 \\
      \end{tabular}
    }
    \label{tab:streak-geometry}
    \vspace{-5pt}
\end{table}

\noindent\textbf{Quantitative Results on Two-stage Strategy.}
% 二阶段法
We further evaluate representative two-stage strategies that first derains individual view with CSUD~\cite{dong2025csud} and then reconstructs with MVSplat~\cite{chen2024mvsplat} or 3DGS~\cite{cite:3dgs}.
Tab.~\ref{tab:twostage} demonstrates that DerainSplat jointly models weather and geometry, outperforming two-stage strategy in both quality and real-time inference.

\subsection{Ablation Study}
We conduct a ablation study of DerainSplat on Rainy RealEstate10K, retaining the same weather-factor prediction.

\noindent\textbf{Core components of DerainSplat.}
Geometry Support guides depth-candidate matching and Gaussian opacity with scene reliability, suppressing weather-induced spurious Gaussian.
Radiance Support denotes support-weighted depth-aligned apearance fusion, restoring corrupted reference colors before Gaussian prediction.
The experimental results are summarized in Tab.~\ref{tab:ablation}, demonstrating the pivotal roles of these modules in rainy--to--clean reconstruction.

\begin{figure}
    \centering
    \vspace{-7pt}
    \includegraphics[width=1\linewidth]{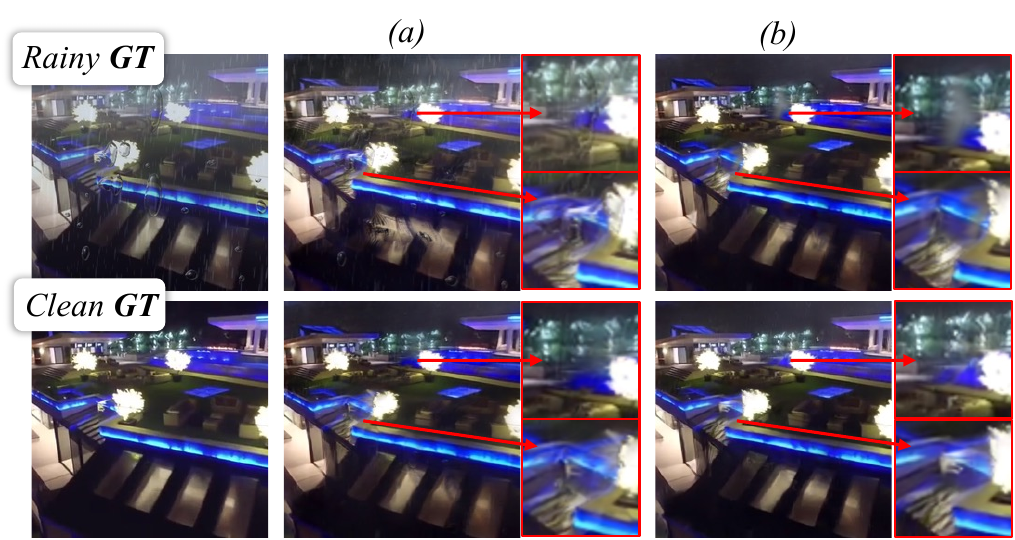}
    \vspace{-13pt}
    \caption{\textit{Comparison of rainy-cycle consistency.} The upper and lower rows are produced without and with the cycle loss, respectively; (a) and (b) trained 1K and 10K steps.}
    \label{fig:ablation-qualitative}
    \vspace{-14pt}
\end{figure}

\noindent\textbf{Importance of Rainy-cycle Consistency.}
Rainy-cycle consistency re-applies the predicted factors to clean context renderings and aligns them with the rainy inputs.
Removing this constraint degrades reconstruction quality, as shown in Tab.~\ref{tab:ablation}~(row 4).
Fig.~\ref{fig:ablation-qualitative} further shows that removing the cycle ratains weather artifacts and slows convergence.

\begin{table}
    \centering
    \renewcommand\arraystretch{1.25}
    \vspace{-5pt}
    \caption{Ablation study on rainy RealEstate10K.}
    \vspace{-5pt}
    \label{tab:ablation}
    \resizebox{1\linewidth}{!}{%
      \begin{tabular}{cc|ccc}
          \textit{Geo. Support} & \textit{Radi.
          Support}
          & PSNR$\uparrow$ & SSIM$\uparrow$ & LPIPS$
          \downarrow$ \\
          \noalign{\hrule height 1pt}
          \xmark & \xmark & 19.73 & 0.726 & 0.304 \\
          \cmark & \xmark & 21.58 & 0.757 & 0.215 \\
          \cmark & \cmark & \textbf{22.77} & \textbf{0.773} & \textbf{0.188} \\
          \noalign{\hrule height 1pt}
          \multicolumn{2}{c|}{w/o rainy-cycle
          consistency}
          & 21.62 & 0.746 & 0.223 \\
      \end{tabular}
    }
\end{table}

%% file: sections/5conclusion.tex
% !TEX root = ../AnonymousSubmission2027.tex
\section{Conclusion}
In this paper, we propose \textbf{\textit{DerainSplat}}, a weather-aware feed-forward framework for reconstructing clean 3D scenes from sparse rainy views.
Our method initially constructs a large-scale multi-view derain dataset synthesized through a four-stage weather formation pipeline, providing paired clean-rain images and privileged factors.
Rather than deraining individual view, DerainSplat disentangles it into explicit factors that jointly guide cross-view geometry matching and depth-aligned appearance fusion within the feed-forward 3DGS pipeline.
Extensive experiments on multiple rainy datasets demonstrate that DerainSplat outperforms existing methods.